\documentclass[10pt,twocolumn,letterpaper]{article}

\usepackage{iccv}
\usepackage{times}
\usepackage{epsfig}
\usepackage{graphicx}
\usepackage{amsmath}
\usepackage{amssymb}

\usepackage[breaklinks=true,bookmarks=false]{hyperref}

\iccvfinalcopy 

\def\iccvPaperID{****} 

\ificcvfinal\fi

\begin{document}

\title{Brand Label Albedo Extraction of eCommerce Products using Generative Adversarial Network}

\author{Suman Sapkota*, Manish Juneja*, Laurynas Keleras*,  Pranav Kotwal*, Binod Bhattarai*!\\ *Zeg.AI  Ltd  London, UK\\
!University College London, UK \\
{\tt\small \{Suman, Manish, Laurynas, Pranav, Binod\}@zeg.ai}
}

\maketitle
\ificcvfinal\thispagestyle{empty}\fi

\begin{abstract}
In this paper we present our solution to extract albedo of branded labels for
e-commerce products. To this end, we generate a large-scale photo-realistic synthetic data set for albedo extraction followed by training a generative model to translate images with diverse lighting conditions to albedo. We performed an extensive evaluation to test the generalisation of our method to in-the-wild images. From the experimental results, we observe that our solution generalises well compared to the existing method both in the unseen rendered images as well as in the wild image. Our data set is publicly available for research purpose~\footnote{shorturl.at/novES}. 
\end{abstract}

\begin{figure*}
    \centering
    \includegraphics[width=0.30\textwidth]{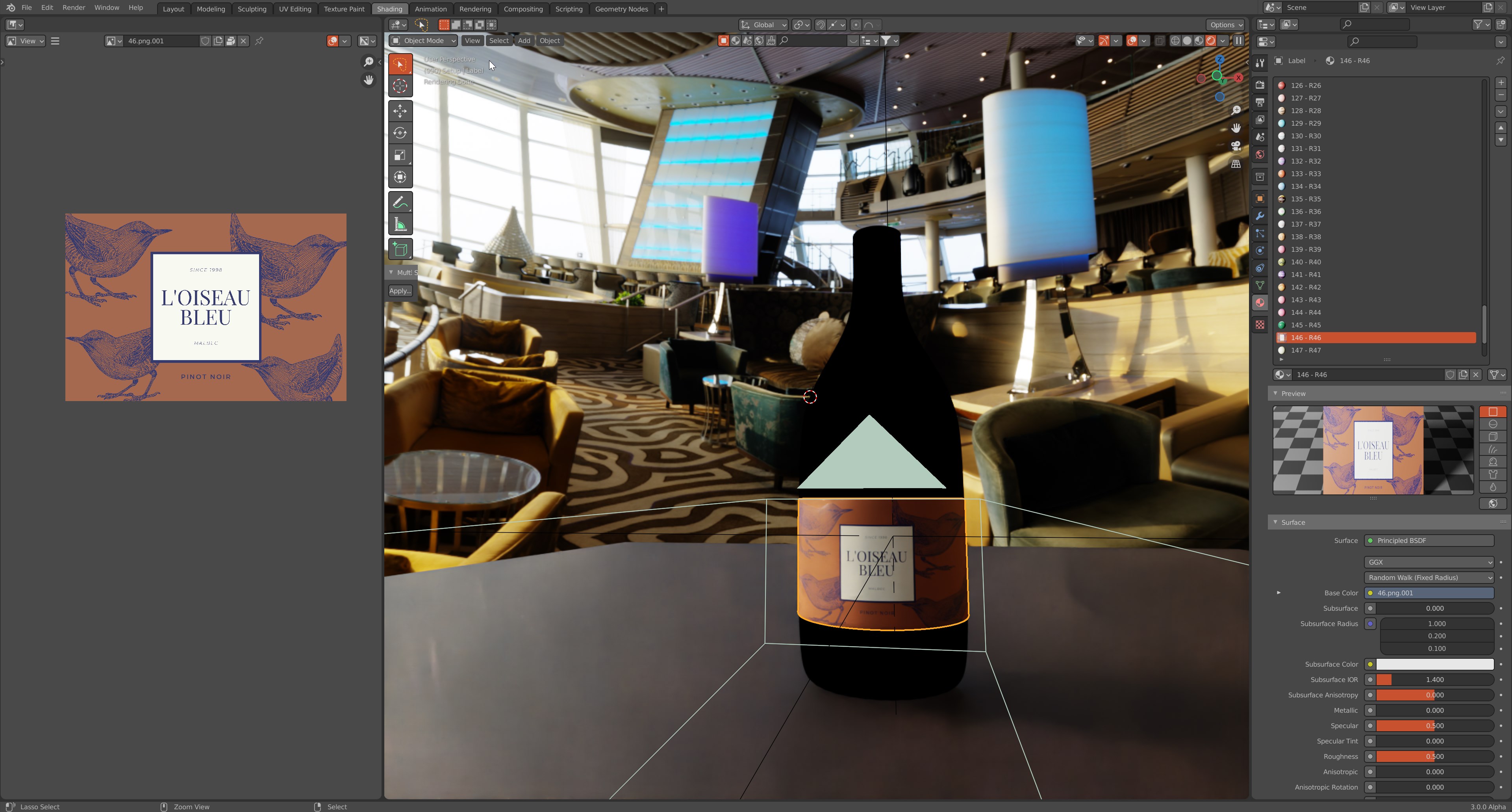}
    \includegraphics[width=0.30\textwidth]{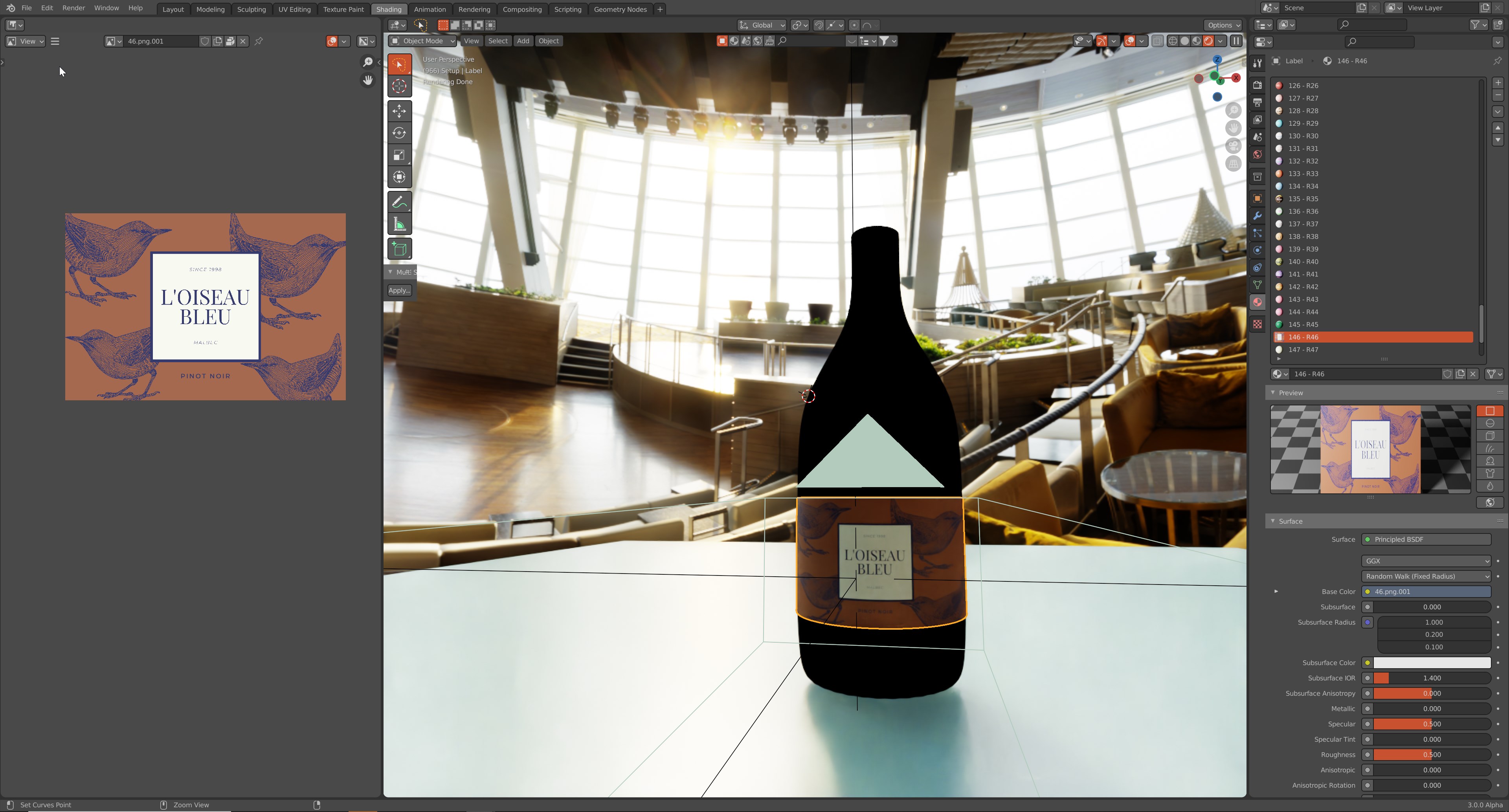}
    \includegraphics[width=0.30\textwidth]{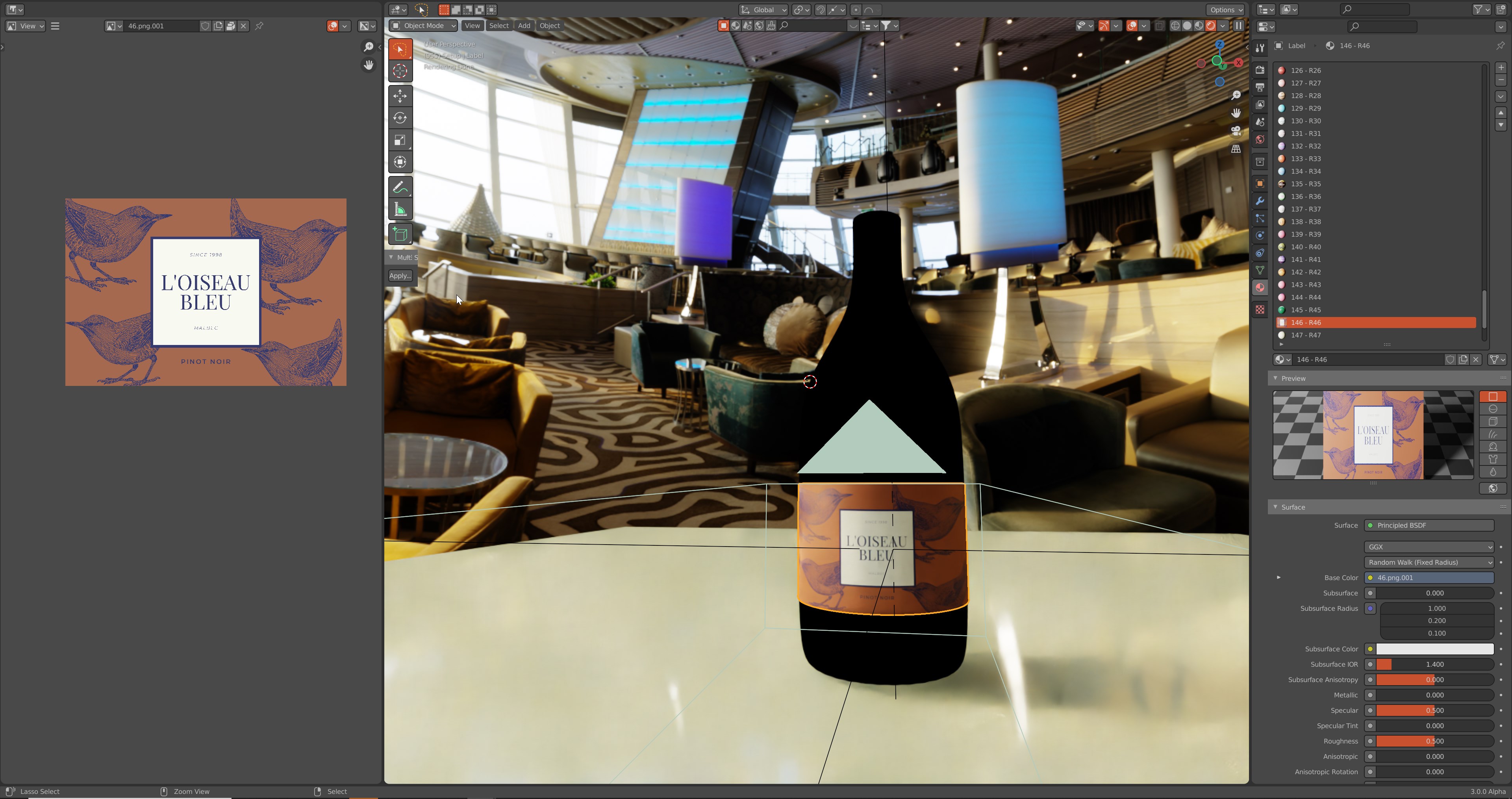}
    \caption{Screenshots showing simulation of natural lighting effect in Blender}
    \label{fig:data_generation_blender}
\end{figure*}

\section{Introduction and Related Works}
Studio Photo-shoots have widely been used to promote e-commerce products in social media and other digital advertising channels for a long time. Product sales are directly correlated with the quality of the photography used to promote a product~\cite{im2021effects}. Shooting a studio level
photography is quite expensive for a small scale brand. In addition to the
cost, there is a massive carbon footprint in shipping products and props to a studio.
In order to address these problems, rendering studio level photography from an
RGB image captured from a smartphone is getting popular.

To summarise on a higher level, rendering a studio level photo from a smartphone-captured photo involves extracting 3D parameters and the branded label from RGB images followed by
slaping the label on the 3D model, and rendering it. 
As we know, the image captured by the smartphone camera sensor is exposed to various intensities of light which directly impacts
the visual perception of the branded labels and logos.  Hence, it is important to correct lighting before slapping the logos on the 3d model and rendering it.
Extracting 3D model parameters from an RGB image is an active research problem~\cite{jin20203d,wang2018pixel2mesh,geiger2011stereoscan}. In this paper, we present our solution to extract albedo map of the branded labels of 
e-commerce products. We have framed this problem as lighting correction of 2D images~\cite{lettry2018darn}. Previous work on lighting correction on 2D images
are focused on external environment and scene~\cite{lettry2018darn} to remove the shades. To the best of our knowledge, this is the first work to extract albedo 
from wide range of e-commerce products labels.

Our contributions are in two folds. First we create a large-scale data set for albedo extraction. We collected a large number of branded labels from the web, prepossessed them carefully in order to remove the unwanted lighting effects using Canva~\footnote{https://www.canva.com/photo-editor/}, an online photo-editing tool. These images act as a ground truth. We load these images to Blender~\footnote{https://www.blender.org/}, and apply different environmental conditions in order to simulate the natural lighting condition. These environmental conditions include mostly of interiors like halls, rooms and closed garages. We render such images and pair them with the ground truth to create training examples. We use these training examples to learn the parameters of
a translator which takes a brand label image with various natural lighting  to translate it to albedo. We train the framework in an adversarial manner similar to that of pix2pix~\cite{isola2017image}. 


One of the closest methods to our work is from Li et al.~\cite{li2018learning}. They also tackled the problem of albedo extraction
from 2D images by generating a large-scale SVBRDF data set followed by regressing a multi-stage deep network. Their method is based 
on Inverse Rendering Deep Models \cite{li2018learning} and demands ground truth of depth maps, surface normals in addition to albedo. Compared to their method, our framework requires only albedo ground-truth and is a single stage-pipeline based on Generative Adversarial Network (GAN). ~\cite{lettry2018darn} is another work using GAN to correct the lighting. However, this method is 
limited to removing only shades and applied on external environment. 

\section{Our Approach}
In this Section, we describe our data (\textbf{ZegLit}) generation pipeline followed by our learning pipeline.

\subsection{Data Collection}
The data was created using DCC software Blender along with Microsoft Paint, Adobe Photoshop and Canva online.

For lighting, the dataset generator consisted of 50 HDRi panorama maps in order to simulate realistic lighting conditions. These panoramas were gathered from HDRI Haven online.  Most of the HDRi maps used simulate lighting conditions of interiors of a building or home. This included extremely well-lit rooms halls and garages where light may be coming from lamps, ceiling light, bulbs or windows. Examples of HDRi were also included where the source of light in the HDRi was very dim including a fireplace at night time or light coming through a window on a gloomy day in the evening. Some other artificial sources of light like customized point lights and area lights were used in the rendering environment besides the HDRi maps to simulate the desired lighting conditions. All of these light sources were created and used inside the Blender environment only.

The images of the label were produced and collected with the help of Paint, Photoshop and Canva. Parts of the label were created in Paint and Photoshop and then brought together in Canva for sharing, assembling and quick manipulation. These images were pure albedo/colour maps at this stage. A wide variety of colours were used to create these images in order not to introduce any bias in the final rendering. After finishing the export from Canva, the label images were introduced inside Blender.  Material properties such as glossiness and metalness were added when these images were introduced inside Blender in order to simulate materials like plastic, cardboard etc. These labels were then UV wrapped around 3D objects. The 3D objects were created all inside Blender and were quite simple. They did not have any complex material properties as they were not to be included in the final rendered image. The final rendered images only have labels visible.

After UV wrapping the labels with proper materials on the 3D objects, setting up appropriate HDRi inside Blender and setting the extra sources of light if necessary; rendering commenced using an orthographic camera. The choice of Orthographic camera was due to its flat look in the final render. Rendering was done using the latest Cycles X rendering engine provided inside Blender. Over 150 Labels were used to render 2 batches of 5000 images. For each new render, random HDRi was assigned. These HDRi were also rotated randomly in order to get as many lighting angles as that particular HDRi could provide. Labels were also procedurally assigned for each new render. Some material properties of labels like roughness were changed procedurally for each render based on a certain threshold.

Figure~\ref{fig:data_generation_blender} summarises the data generation pipeline. We collected in total 5000 labels of consumer goods with wide range of variations in lighting and material properties. 

\begin{figure*}
    \centering
    \includegraphics[trim=0cm 0cm 5cm 4cm, clip, width=0.75\linewidth]{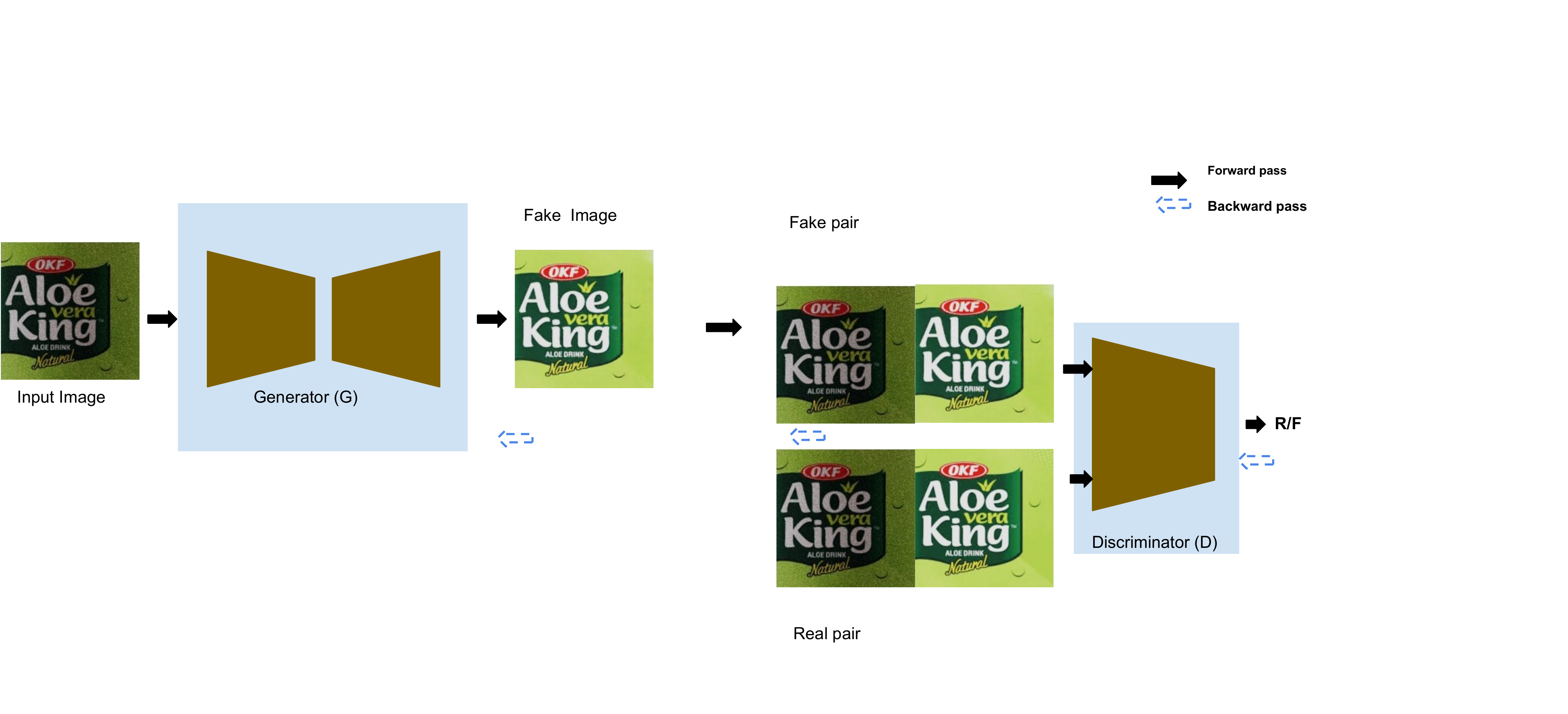}
    \caption{Schematic diagram showing the proposed pipeline. We feed in the label with various lighting to the generator. The generator learns the parameters to translate the input image to their corresponding albedo. The translated image along with the input image make a pair called "fake pair". Similarly, the input image and its ground truth make another called "real pair". We feed both these pairs to the discriminator.  We learn the parameters of the generator and discriminator in an adversarial manner. 
    The generator learns the parameters to puzzle discriminator whereas the discriminator learns to better distinguish fake pairs from real pairs. }
    \label{fig:proposed_pipeline}
\end{figure*}
\subsection{Learning from Data}
After the seminal work on Generative Adversarial Network (GAN)~\cite{NIPS2014_5ca3e9b1}, it has been successfully applied to several tasks including image editing~\cite{choi2018stargan}, style transfers~\cite{karras2019style},
domain adaptation~\cite{gecer2018semi}, etc. In this paper, we present a GAN pipeline based on pix2pix~\cite{isola2017image} network to extract albedo of the wide ranges of e-commerce product labels. 

Figure~\ref{fig:proposed_pipeline} details the proposed pipeline. In the Figure,  \textbf{G} represents the generator and  \textbf{D} represents the discriminator. The generator
is implemented by an encoder and a decoder with an architecture similar to that of Unet~\cite{ronneberger2015u} and we implement discriminator by a convolutional neural network. For an 
input product label $I_x$ extracted from a product image captured in a natural lighting environment is first fed into the generator. The generator learns the 
parameters to extract the albedo from the input label and return $I_x^f$. The output from the generator and the input image makes a pair ($I_x, I_x^f$) that we call a fake pair.
Similarly, another pair with an input image with its corresponding ground-truth ($I_x, I_x^g$) makes a positive pair.  $I_x^g$ is a ground truth albedo image. We set the label of real pair as 1 and that of fake pair as 0. We learnt the parameters of both the generator and the discriminator to optimise the objective as given in Equation~\ref{eqn:main_equation} in an adversarial manner. The generator learns the parameters to make the image as realistic as the ground truth albedo whereas the discriminator learns the parameters to discriminate real vs fake pairs correctly. 

\begin{equation}
\begin{split}
    \mathcal{L}_{adv} = \mathop{\mathbb{E}}_{I_x^r, I_g^r \sim I^r}[\log(D_{adv}(G(I_x^r,I_g^r)))] - \\ 
    \mathop{\mathbb{E}}_{I_x^r \sim I^r, I_x^f \sim I^f}[\log(1- D_{adv}(I^r_x, I_x^f))] 
\end{split}
\label{eqn:main_equation}
\end{equation}

\section{Experiments}
We evaluated our method on two benchmarks and performed extensive qualitative analysis to validate our idea.  

\textbf{ZegLit Dataset:}  This data set consists of 5,000 images with 150 different brand labels. 
We split the data set into 80\% as train set and 20\% as test set. We train the model on the training set and evaluate on test set. 


\textbf{SVBRDF Dataset~\cite{li2018learning}:} This data set consists of 200K images with  albedo, specular roughnes, surface normal, and depth. For comparison, we only take albedo into consideration. 

\subsection{Qualitative Evaluations:} We performed extensive qualitative comparisons to demonstrate effectiveness of the proposed pipeline.

\begin{figure}
    \centering
    \includegraphics[width=0.32\linewidth]{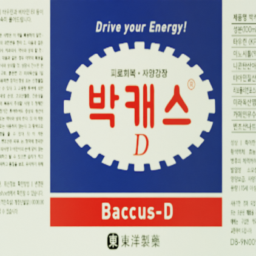}
    \includegraphics[width=0.32\linewidth]{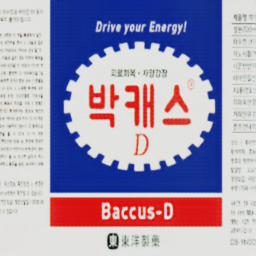}
    \includegraphics[width=0.32\linewidth]{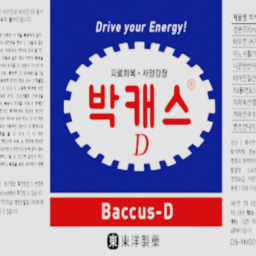}
     
    \includegraphics[width=0.32\linewidth]{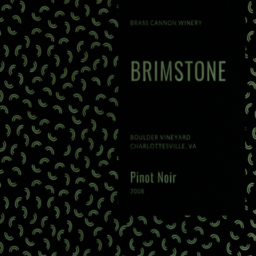}
    \includegraphics[width=0.32\linewidth]{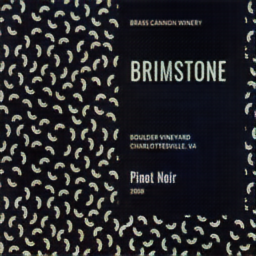}
    \includegraphics[width=0.32\linewidth]{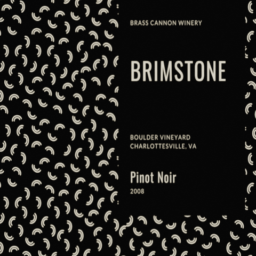}
    
    \includegraphics[width=0.32\linewidth]{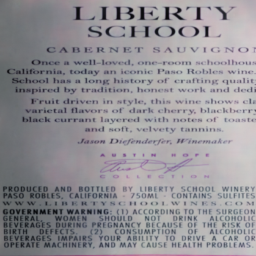}
    \includegraphics[width=0.32\linewidth]{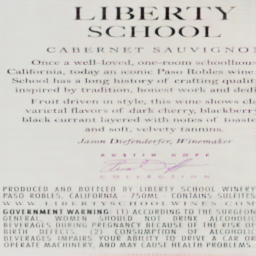}
    \includegraphics[width=0.32\linewidth]{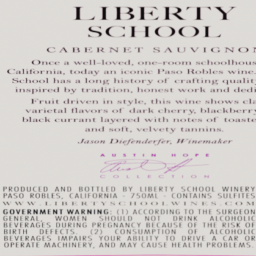}

   \includegraphics[width=0.32\linewidth]{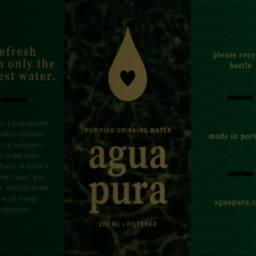}
   \includegraphics[width=0.32\linewidth]{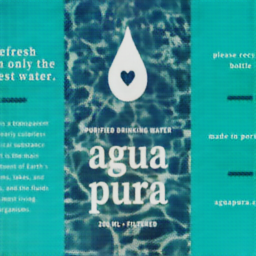}
   \includegraphics[width=0.32\linewidth]{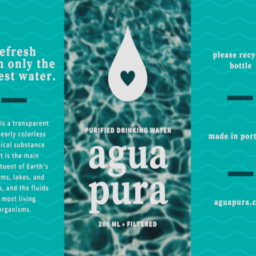}

   \includegraphics[width=0.32\linewidth]{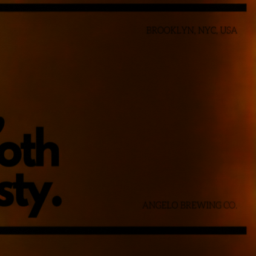}
   \includegraphics[width=0.32\linewidth]{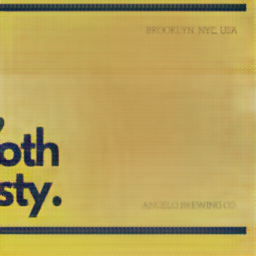}
   \includegraphics[width=0.32\linewidth]{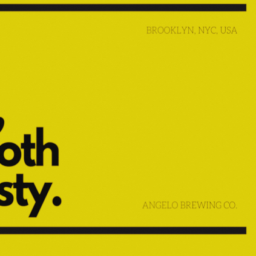}

   \includegraphics[width=0.32\linewidth]{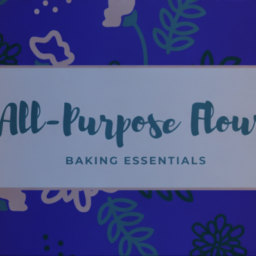}
   \includegraphics[width=0.32\linewidth]{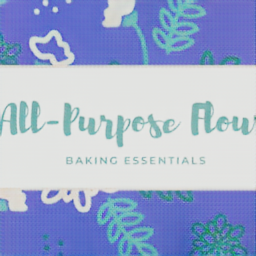}
   \includegraphics[width=0.32\linewidth]{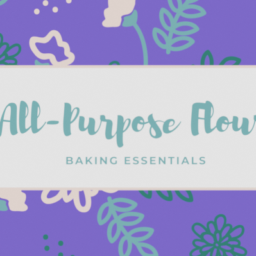}
  
   \includegraphics[width=0.32\linewidth]{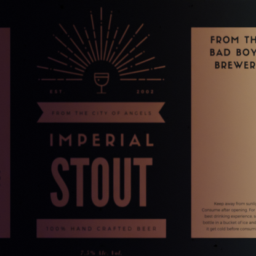}
   \includegraphics[width=0.32\linewidth]{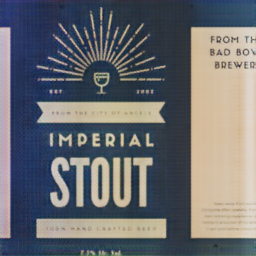}
   \includegraphics[width=0.32\linewidth]{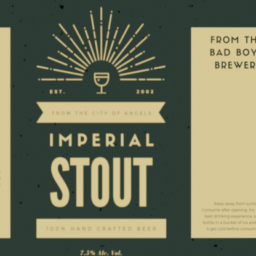}
     
    \caption{Qualitative results on test examples on Zeg brand label data set. \textbf{Left} column shows the input images, \textbf{middle} column shows the albedo extracted by
    our model and the \textbf{right} column shows the ground-truths. From these examples we can see that our method generalises to the unseen examples. }
    \label{fig:test_set}
\end{figure}

\textbf{Generalisation to unseen labels:} Figure~\ref{fig:test_set} shows the performance of our framework trained on ZegLit data set. In the Figure, the first column shows some of the randomly selected test  labels. The middle column is the albedo extracted by our model. Whereas, the third column is the ground truth. From the comparisons, we can conclude that our method can generalise well with various types of labels exposed with uncontrolled lighting conditions.

\begin{figure}
    \includegraphics[width=0.24\linewidth]{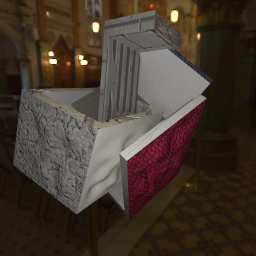}
    \includegraphics[width=0.24\linewidth]{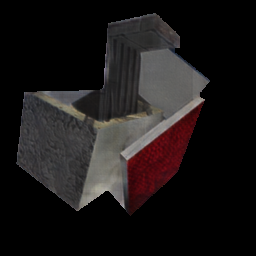}
    \includegraphics[width=0.24\linewidth]{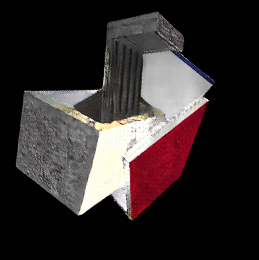}
    \includegraphics[width=0.24\linewidth]{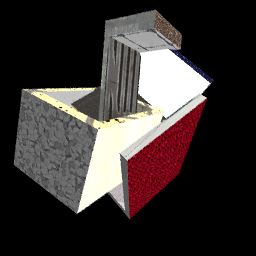}
    
    \includegraphics[width=0.24\linewidth]{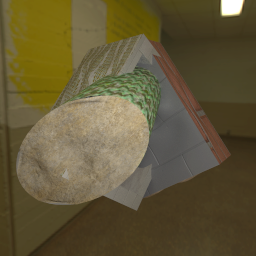}
    \includegraphics[width=0.24\linewidth]{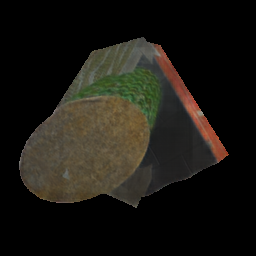}
    \includegraphics[width=0.24\linewidth]{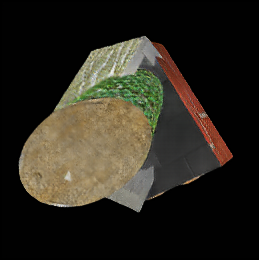}
    \includegraphics[width=0.24\linewidth]{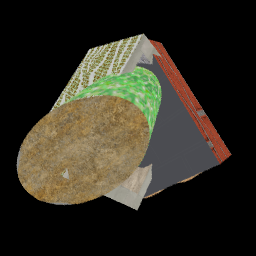}

    \includegraphics[width=0.24\linewidth]{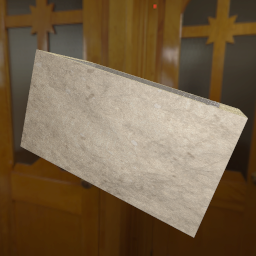}
    \includegraphics[width=0.24\linewidth]{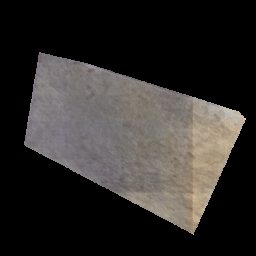}
    \includegraphics[width=0.24\linewidth]{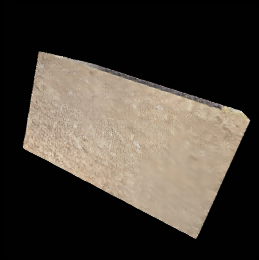}
    \includegraphics[width=0.24\linewidth]{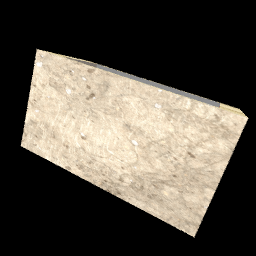}
    \caption{ Comparison with the existing art~\cite{li2018learning}. From left to right: Input, Output of ~\cite{li2018learning}, Ours, and ground truth}
    \label{fig:comparison_prev_art}
\end{figure}

\textbf{Comparison to previous method:}
We compare the performance of our method with one of the recent works on inverse rendering~\cite{li2018learning} which learns the parameters of multi-head multi-stage Convolutional Neural Network to regress shape, albedo, and relflectance jointly. Compared to them~\cite{li2018learning}, we only rely on albedo and our pipeline is single stage.
Thus, our method needs less annotations to learn the parameters and also less hassle to train being a single stage pipeline. Figure~\ref{fig:comparison_prev_art} shows the performance comparison on SVBRDF data set. In the Figure, from left to right, each column represents input image, albedo by ~\cite{li2018learning}, ours and the ground truth respectively. From the qualitative results, although our pipeline only utilises albedo as ground truth information, it still manages to obtain the better performance compared to ~\cite{li2018learning}.  

\begin{figure}
    \centering
    \includegraphics[width=0.32\linewidth]{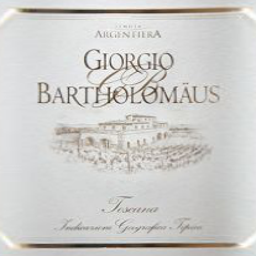}
   \includegraphics[width=0.32\linewidth]{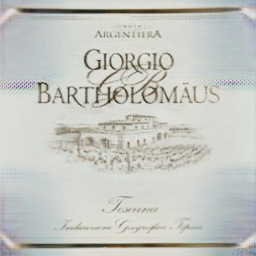}
   \includegraphics[width=0.32\linewidth]{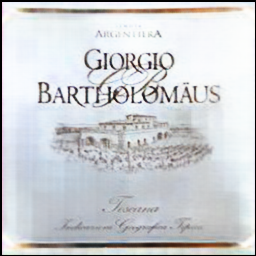}
   
   \includegraphics[width=0.32\linewidth]{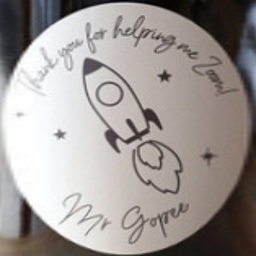}
   \includegraphics[width=0.32\linewidth]{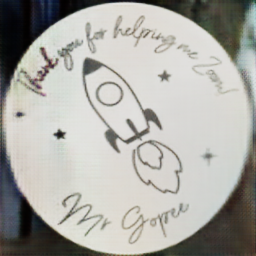}
   \includegraphics[width=0.32\linewidth]{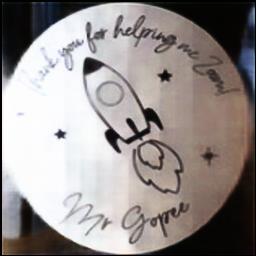}
 
   \includegraphics[width=0.32\linewidth]{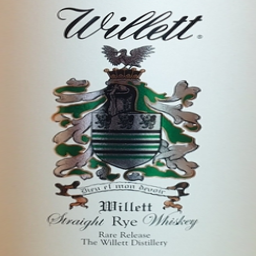}
   \includegraphics[width=0.32\linewidth]{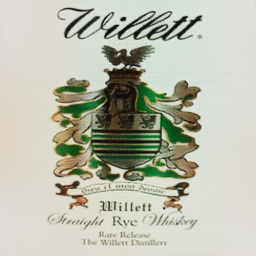}
   \includegraphics[width=0.32\linewidth]{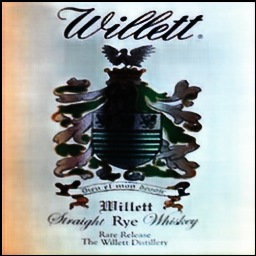}
   
   \includegraphics[width=0.32\linewidth]{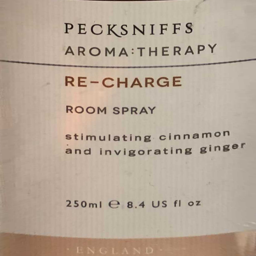}
   \includegraphics[width=0.32\linewidth]{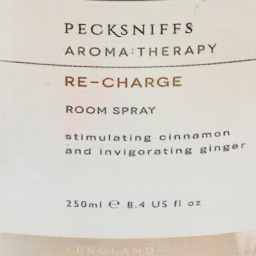}
   \includegraphics[width=0.32\linewidth]{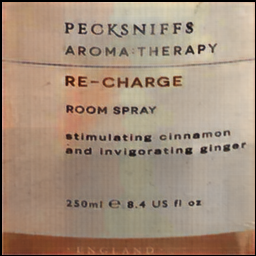}
 
   \includegraphics[width=0.32\linewidth]{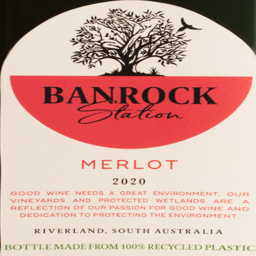}
   \includegraphics[width=0.32\linewidth]{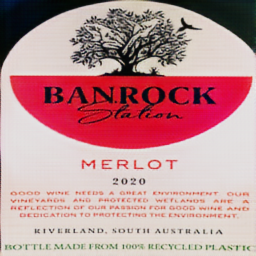}
   \includegraphics[width=0.32\linewidth]{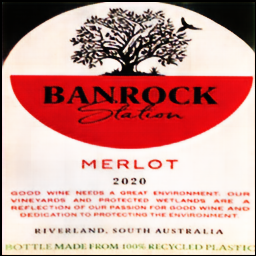}
 
    \caption{Qualitative comparison between our method  and ~\cite{li2018learning} on natural images. First column is input, second is from our model trained on our data set and the third column shows the output from~\cite{li2018learning}}
    \label{fig:compare_wild_images}
\end{figure}

\textbf{Generalisation to real data:} We capture few pictures of different wine bottles by iPhone 11 and fed to a pre-processing pipeline to extract the labels from the 2D images. First column on Figure~\ref{fig:compare_wild_images} shows the labels we extracted. These labels were fed to our model trained on Zeglit data and a pre-trained model from ~\cite{li2018learning}. Middle Column shows the albedo extracted by our model whereas the last column is by ~\cite{li2018learning}. From these comparisons, we can observe that our model generalises well to the images captured by the commonly used smartphone such as iPhone 11.  

\section{Conclusions}
In this paper, we present a pipeline for albedo extraction of branded labels for eCommerce products. To this end, we made two contributions.One, we collected a large scale photo-realistic data set for albedo extraction. On the other hand, we train an albedo extraction pipeline using pix2pix generative adversarial network. From the extensive qualitative experiments on two different data sets, we demonstrate the collected data set can generalise well on both unseen examples as well as the natural images. Also, GAN based single stage model is more effective than a multi-stage pipeline. 


\clearpage

{\small
\bibliographystyle{ieee_fullname}
\bibliography{egbib}
}

\end{document}